\documentclass[letterpaper, 10pt, conference]{ieeeconf} 
\IEEEoverridecommandlockouts                              %

\makeatletter
\newcommand*\titleheader[1]{\gdef\@titleheader{#1}}
\AtBeginDocument{%
  \let\st@red@title\@title
  \def\@title{%
    \bgroup\normalfont\large\centering\@titleheader\par\egroup
    \vskip1.5em\st@red@title}
}
\makeatother

\usepackage{graphicx} %
\usepackage{mathptmx} %
\usepackage{amsmath} %
\usepackage{amssymb}  %
\usepackage{multirow}                     %
\usepackage{cite}
\usepackage[size=footnotesize]{todonotes}

\title{\LARGE \bf Study on Soft Robotic Pinniped Locomotion}
\titleheader{\small{This paper has been accepted to 2023 IEEE/ASME International Conference on Advanced Intelligent Mechatronics (AIM).}}

\author{Dimuthu~D. K.~Arachchige$^{1}$, Tanmay Varshney$^{2}$, Umer Huzaifa$^{1}$, Iyad Kanj$^{1}$, Thrishantha Nanayakkara$^{3}$, \\ Yue Chen$^{4}$, Hunter~B.~Gilbert$^{5}$, and Isuru~S.~Godage$^{6}$%
	\thanks{$\!\!\!\!\!\!\!\!\!\!^{1}$School of Computing, Jarvis College of Computing and Digital Media, DePaul University, Chicago, IL 60604, USA.\,
		\newline Corresponding author: {\tt\small DARACHCH@depaul.edu}\,
        \newline
        $^{2}$College of Engineering, Ohio State University, Columbus, OH 43210, USA.
		$^{3}$Dyson School of Design Engineering, Faculty of Engineering, Imperial College London SW7 2BX, UK.
		$^{4}$Department of Biomedical Engineering, Georgia Institute of Technology, Atlanta, GA 30332, USA.		
		$^{5}$Department of Mechanical and Industrial Engineering, Louisiana State University, Baton Rouge, LA 70803, USA.
		$^{6}$Department of Engineering Technology and Industrial Distribution, Texas A\&M University, College Station, TX 77843, USA.
	}
}

\begin{document}
	\maketitle
	\thispagestyle{empty}
	\pagestyle{empty}
	
	\begin{abstract}
		
Legged locomotion is a highly promising but under--researched subfield within the field of soft robotics. The compliant limbs of soft-limbed robots offer numerous benefits, including the ability to regulate impacts, tolerate falls, and navigate through tight spaces. These robots have the potential to be used for various applications, such as search and rescue, inspection, surveillance, and more. 
%Despite the potential of soft-limbed robots, the c
%Current 
The state-of-the-art still faces many challenges, including limited degrees of freedom, a lack of diversity in gait trajectories, insufficient limb dexterity, and limited payload capabilities.
To address these challenges, 
%our research focuses on developing 
we develop a modular soft-limbed robot that can mimic the locomotion of pinnipeds. By using a modular design approach, we aim to create a robot that has improved degrees of freedom, gait trajectory diversity, limb dexterity, and payload capabilities. 
We derive a complete floating-base kinematic model of the proposed robot and use it to generate and experimentally validate a variety of locomotion gaits. Results show that the proposed robot is capable of replicating these gaits effectively. We compare the locomotion trajectories under different gait parameters against our modeling results to demonstrate the validity of our proposed gait models.

	\end{abstract}
	
	\section{Introduction\label{sec:Introduction}}
	
%	\todo[inline]{Minor comment: Change math font types: scalars - Italics (lower or uppercase), vectors - bolt italics (lower or uppercase), Matrices - Uppercase bold}
	
	Soft robots are manufactured with flexible materials (e.g., elastomers, fabrics, polymers, shape memory alloy) and are mostly actuated through pneumatic and hydraulic pressure, tendons, and smart materials \cite{rus2015design}. Soft mobile robots -- a branch of the soft robot family -- use compliant structures (e.g., body, limbs, etc.) to achieve locomotion. They are mostly designed to mimic the behavior (typically locomotive patterns) of biological creatures \cite{ahmed2022decade}.
	Compared to rigid mobile robots, the inherently compliant elements
    in soft mobile robots enable them to absorb ground impact forces without active impedance control \cite{al2018active}.
	Furthermore, their ability to deform actively and passively
	allows them to
	gain access to confined areas \cite{hawkes2017soft}.
    As a result, 
	soft mobile robots have great potential to replace humans in performing dangerous 
	tasks, such as nuclear site inspection \cite{oshiro2018soft}, search and rescue operations \cite{talas2020design}, and planetary exploration \cite{ng2021untethered}.

	\begin{figure}[tb] 
		\centering
		\includegraphics[width=.99\linewidth]{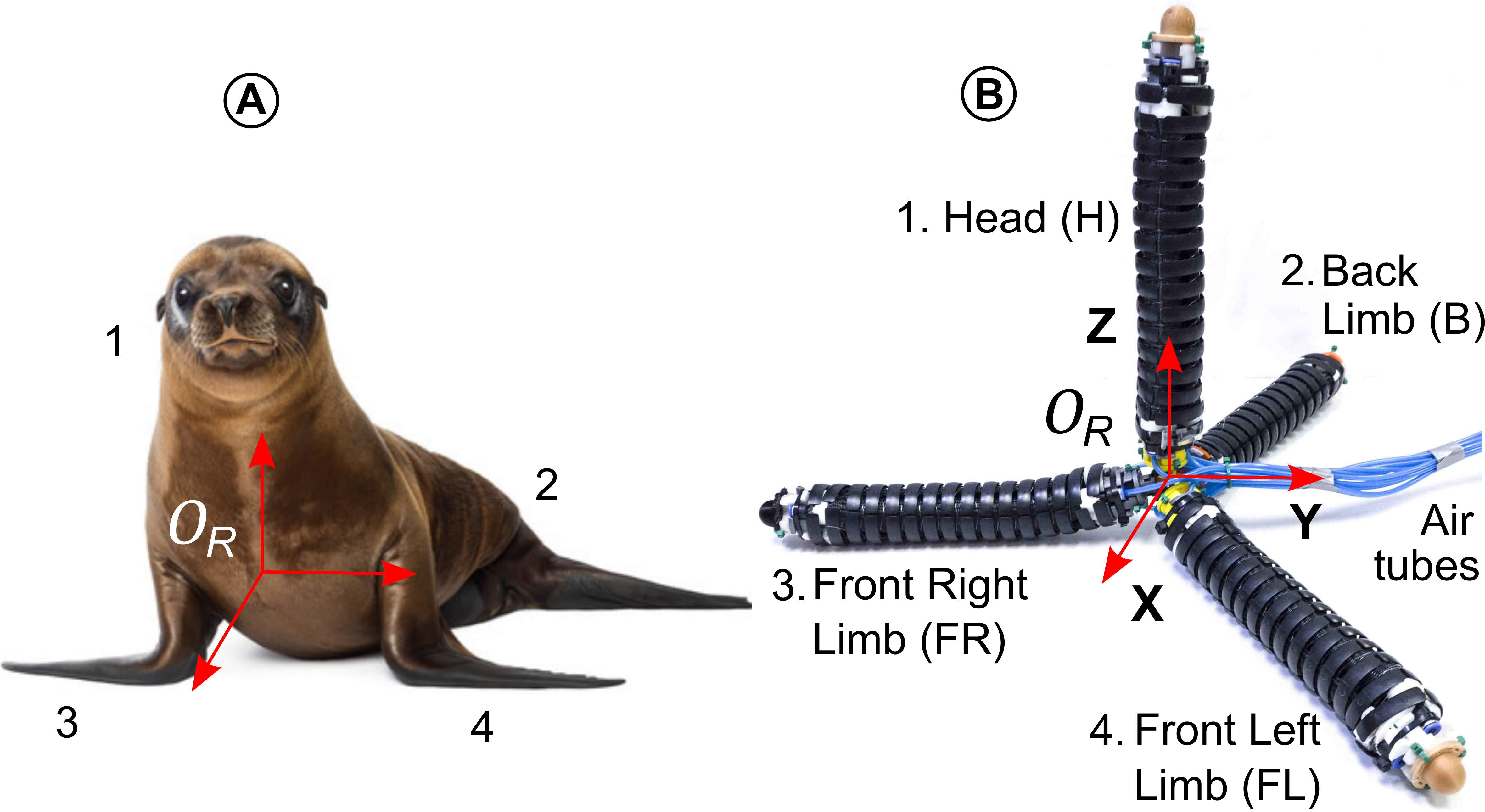}
		\caption{(A) Bioinspiration from pinnipeds (i.e., seals, sea lions, and walruses) that use fore flippers and body (or hind flipper) for terrestrial locomotion. (B) Pinniped robot in an unactuated pose.}
		\label{fig:Fig1_IntroductionImage} 
	\end{figure}
	
%	As potential solutions in these application areas, m
	Many soft-limbed robotic prototypes have been proposed to date \cite{sun2021soft}.
%    Their designs often suffer a lack of degrees of freedom (DoF), limb dexterity, and payload capabilities. 
%	\todo{Iyad: Could you review the related literature in a story-telling way instead of listing them in this abrupt fashion?}
	For instance, the pneumatically actuated multi-gait robot reported in \cite{shepherd2011multigait} uses 
%	only 
	five actuators 
%	and a simple pneumatic valving system that operates at low pressures. A 
	to generate 
%	combination of 
	crawling and undulation gaits. 
%	allows the robots to 
%	navigate obstacles. 
	However, it was only capable of preprogrammed straight locomotion 
	without turning. 
%	
%	while t
	The autonomous untethered quadruped 
	in \cite{tolley2014resilient} is 
%	fabricated using composite ms of silicone elastomer, polyaramid fabric, and hollow glass microspheres, and 
%	strong enough to 
	capable of carrying the subsystems (i.e., miniature air compressors, a battery, valves, and a controller). 
	The robot can operate under adverse environmental conditions but only supports limited gaits. 
	The quadruped in \cite{godage2012locomotion} can achieve various dynamic locomotion gaits such as crawling and trotting but without turning.
	The quadruped in \cite{drotman2021electronics} presents a new approach for controlling the gaits of soft-legged robots using simple pneumatic circuits without electronic components. The need for preprograming the gaits offer limited gait diversity. 
%	has soft limbs attached to a single point and thus the locomotion is hindered by poor balancing.
%	
	The soft robot prototypes 
	reported in \cite{mao2016locomotion,huang2018chasing}  have stiffness-tunable limbs and are inspired by starfish, including its locomotion and water-vascular systems. 
%	The main limitation thereof are the 
%	suffer from 
	But the 
	low speed and low efficiency 
%	thereof 
	due to 
%	dielectric elastomer 
	shape memory alloy actuators limit their utility.
	The amphibious soft robot 
	in \cite{faudzi2017soft} uses
	highly compliant limbs with no stiffness tunability and resulting in unstable and slow locomotion.
	The soft-limbed hexapod proposed in \cite{suzumori1996elastic} showed the ability to derive a variety of gaits. The hexapod appeared in \cite{liu2020sorx} showed the ability to traverse challenging terrains.
%	 rough, steep, and unstable terrains.
    The large number of limbs however increases the robots' complexity at the cost of limb dexterity. 
%    Typically, robots with more limbs lack limb dexterity.
% 
%    A potential solution to address the above limitations is to consider a tripodal robot  
%    that mimics pinniped locomotion.
%    
    The soft-limbed robot proposed in~\cite{wang2021design} 
%	 a soft-limbed robot in the 
%	has 
	uses only four limbs in spatially symmetric tetrahedral topology. 
%	that can replicate steering, crawling, and rolling gaits. 
%	Thus its design is less complex but capable of replicate 
%	steering, crawling, and rolling gaits. 
%	due to the tripodal assembly. 
	But due to the use of solenoid valves
%	to control the pressure inputs
	-- binary actuation, it 
%	can achieve 
	has limited control of the
	locomotion gaits. In addition, 
	no analytical gait derivation approach 
	was reported and 
%	thus 
	only demonstrated 
	preprogrammed 
%	fundamental 
	locomotions. 
%	The tetrahedral topology presents a spatially symmetric robot design. 
%	Similar to the terrestrial locomotion of pinnipeds (Fig. \ref{fig:Fig1_IntroductionImage}A), two limbs can be employed for crawling while the remaining limbs can be actuated as the body and head
%	\todo{Iyad: ``Can be actuated as the body and head'' sounds awkward. Maybe ``employed'' instead of ``actuated''?} during numerous locomotion modes.  
	
    We propose a new soft-limbed pinniped robot to address the above limitations. 
    We adopt a modular design approach to increase the robustness and utilize hybrid soft limbs with improved payload and dexterous capabilities to fabricate the robot. 
%	Our modular limb design approach increases the robustness of the pinniped assembly. 
	In addition, we present a systematic approach to derive novel locomotion gaits. 
    % Therein, gait modeling is applied to obtain novel gaits.\todo{ISG. Iyad: This sentence sounds empty and noninformative. Maybe be more specific about where the novelty comes from.} 
	Further, we adopt a 
	proportional limb bending mechanism to achieve improved workspace and control. The robot validates locomotion at a 38-fold speed increase than that of the state-of-the-art robot in~\cite{wang2021design}.
%	
%	In this work, o
	Our main technical contributions are: i) designing and fabricating a novel pinniped robot using hybrid soft 
	limbs;
	ii) deriving a complete floating-base kinematic model;
	iii) employing the kinematic model to derive fundamental limb movements; 
	iv) parameterizing 
	limb movements to derive new, sophisticated locomotion gaits; 
	and v) experimentally validating the 
	locomotion gaits under varying gait parameters.

	\begin{figure}[tb] 
		\centering
		\includegraphics[width=1\linewidth]{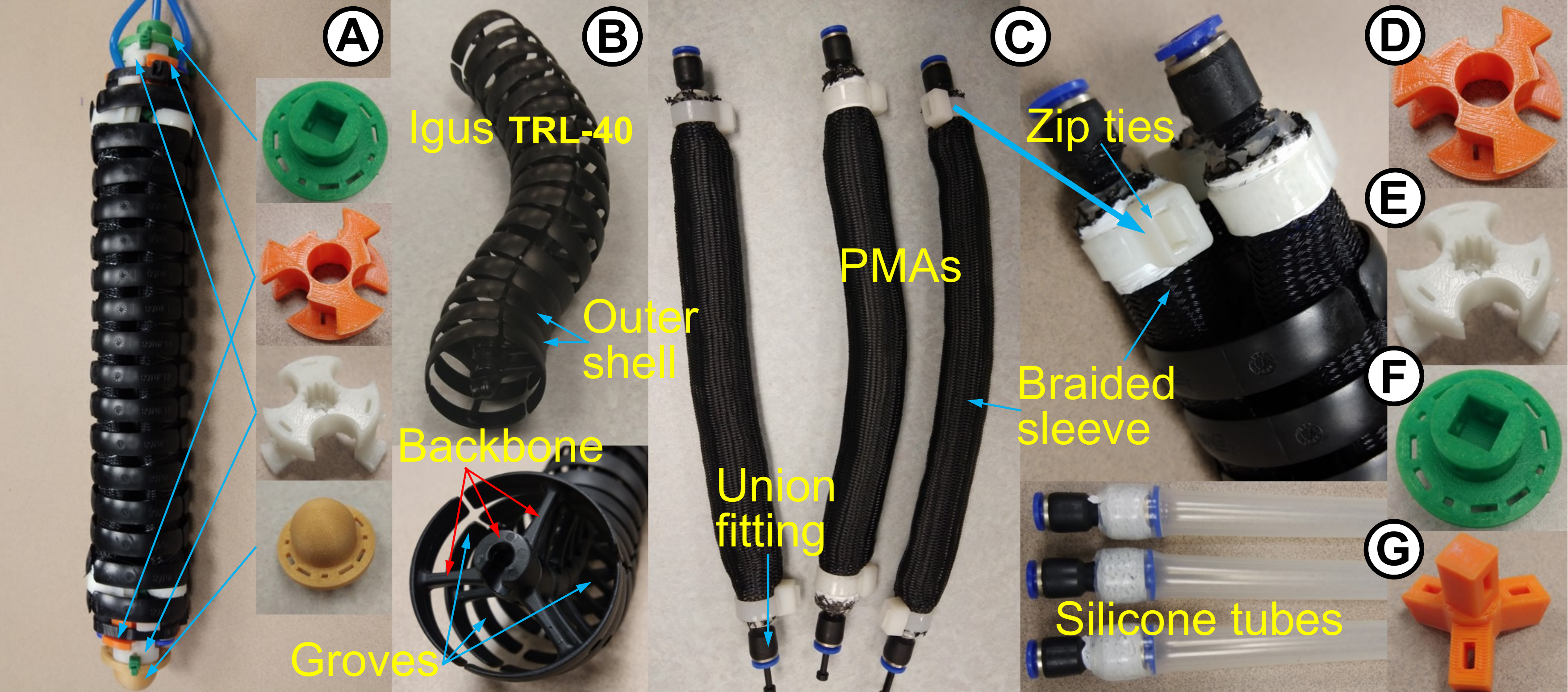}
		\caption{(A) Assembled soft limb. (B) Rigid kinematic chain. (C) PMAs. (D) Edge cap. (E) Middle joint. (F) Upper joint. (G) Tetrahedral joint.}
		\label{fig:Fig2_PartDesign} 
	\end{figure}

	\section{System Model\label{subsec:System-Model}}	
	
	\subsection{Prototype Description\label{sec:Prototype-Description}}

	The 
	proposed soft-limbed pinniped robot 
	is shown in Fig. \ref{fig:Fig1_IntroductionImage}B. It consists of 4 identical soft limbs: Head (H), Back limb (B), Front Right limb (FR), and Front Left limb (FL). 
    %done \todo{capitalize front right/left, back, and head limb references}
	A soft limb 
	(Fig. \ref{fig:Fig2_PartDesign}A)
	is actuated by three
	pneumatic muscle actuators (PMAs) and structurally supported by a 
	backbone and outer shell
	(Fig. \ref{fig:Fig2_PartDesign}B). 
	PMAs are fabricated using silicone tubes (Fig. \ref{fig:Fig2_PartDesign}C). PMAs are inserted into radially symmetric grooves (or channels) of the backbone structure
	(Figs. \ref{fig:Fig2_PartDesign}B and  \ref{fig:Fig2_PartDesign}C)
	and 
	further supported by 3D-printed parts shown in Figs. \ref{fig:Fig2_PartDesign}D, \ref{fig:Fig2_PartDesign}E, and \ref{fig:Fig2_PartDesign}F. 
	We use Nylon threads to wrap PMAs in parallel to the backbone -- in a way that the wrapping does not affect the bending performance -- to prevent buckling upon extension during operation.
	This PMA and backbone arrangement results in an antagonistic 
	actuator configuration since the backbone constrains the length change of PMAs during operation without constraining 
	the omnidirectional bending. 
	Further, 
	the protective shell protects PMAs from potentially damaging environmental contacts.
	A soft limb has an effective length of $240~mm$, a diameter of $40~mm$, and a weight of $0.15~kg$. As shown in our previous work \cite{arachchige2022hybrid, arachchige2021novel,arachchige2023wheelless}, this hybrid design (i.e., combining both soft and hard elements) increases the achievable stiffness range (hence payload) and provides decoupled stiffness and pose control. 
	
    We connect four soft limbs using a 3D-printed tetrahedral joint (Fig. \ref{fig:Fig2_PartDesign}G) to obtain the
	pinniped
	topology (Fig. \ref{fig:Fig1_IntroductionImage}).
	Thus,
	the robot 
	has 12-DoF (3-DoF per limb)
	and weighs $0.65~kg$. In pinnipeds, the bulk of the mass is distributed toward the body (i.e., back end). However, we adopt this topology with symmetric mass distribution to optimize movements in all directions. Further, its spatial symmetry enables reorientation and thus better stability.

	\subsection{Kinematics of Soft Limbs\label{subsub:Forward-Kinematic-Model}}

     \begin{figure}[tb] 
		\centering
		\includegraphics[width=1\linewidth]{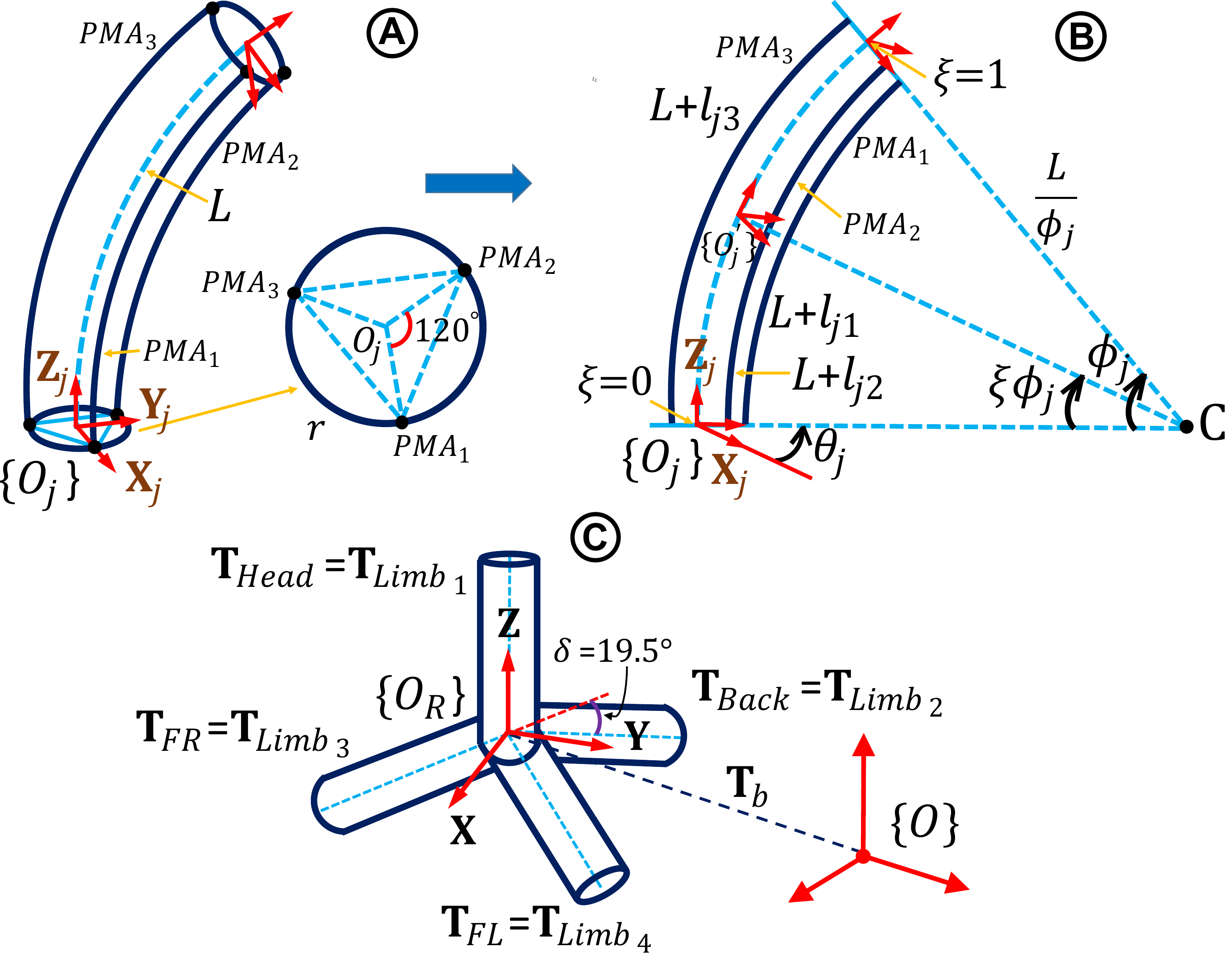}
		\caption{(A) Schematic of a soft limb design with actuator arrangement. (B) A view from an angle normal to the bending plane showing curve parameters. (C) Schematic of the pinniped robot.}
		\label{fig:Fig3_Schematicdiagram} 
	\end{figure}
 
	Figures \ref{fig:Fig3_Schematicdiagram}A and \ref{fig:Fig3_Schematicdiagram}B show the schematic of 
	a soft limb. Consider any \textit{j}-th limb at a given time $t$. Let the length change of PMAs due to actuation be ${l_{ji}(t)} \in {\rm {\mathbb{R}}}$, for $i \in \{ 1,2,3\}$ and $j \in \{ 1,2,3,4\}$ where $i$ and $j$ stand for the PMA number and limb index, respectively. 
	Thus, the joint variable vector of the \textit{j}-th limb 
	is expressed as $\textbf{\textit{q}}_{j} = \left[ {{l_{j1}},\,{l_{j2}},\,{l_{j3}}} \right]^\mathsf{T}$. The time dependency is omitted for brevity.	
	
	The body coordinate system of any \textit{j}-th soft limb, $\{\boldsymbol{O_{j}}\}$,
	is defined at the geometric center of the cross-section on one end (termed base)
	with the first PMA 
	-- associated with 
	$l_{j1}$ joint variable -- anchor point coinciding with the $+X_j$ axis (Fig. \ref{fig:Fig3_Schematicdiagram}A).
	The remaining PMAs with jointspace parameters 
	$l_{j2}$ and $l_{j3}$ are indexed in counterclockwise direction at 
	$\frac{2\pi}{3}$ angle
	offsets about $+Z_j$ from each other at $r_j$ distant from the origin of $\{\boldsymbol{O_{j}}\}$. 
 % origin \todo{Iyad: Do you mean from the origin $O_j$? If so, rewrite it as such.}
	
	When PMAs are actuated with a resulting
	differential pressure,
	the torque imbalance at either end of the soft limb causes it to bend. 
	As is the case with prior work on similarly-arranged soft robots, 
	due to the uniform and symmetric construction, we 
	can approximate 
	that the limb's neutral axis bends in a circular arc. 
	Hence, the spatial pose of a soft limb can be parameterized by the angle subtended by the circular arc, $\phi_{j}$, and the angle to the bending plane w.r.t.~the $+X_j$ axis, $\theta_{j}$. The radius of the circular arc can be derived as $\frac{L}{\phi_{j}}$ where $L\in\mathbb{R}$ is the unactuated length of a PMA (Fig. \ref{fig:Fig3_Schematicdiagram}B). Using basic arc geometry~\cite{godage2015modal},
	the curve parameters are derived using PMA lengths as 
	\begin{align}
		L+l_{ji} & =\left\{ \textstyle \frac{L}{\phi_{j}}-r\cos\left(\textstyle \frac{2\pi}{3}\left(i-1\right)-\theta_{j}\right)\right\} \phi_{j}, \mbox{where} \nonumber \\
		l_{ji}& = -r_j\phi_{j}\cos\left(\textstyle \frac{2\pi}{3}\left(i-1\right)-\theta_{j}\right).    	    	\label{eq:l2cp}
	\end{align}
	
	Since the soft limb is inextensible, the sum of length changes of PMAs (i.e., jointspace variables) for all $i$ in \eqref{eq:l2cp} add up to zero. This results in the length constraint $l_{j1}=-\left(l_{j2}+l_{j3}\right)$, indicating 
	that 
	the complete limb kinematics 
	can be expressed by 
	two independent
	jointspace 
	variables. From \eqref{eq:l2cp} and following \cite{godage2015modal,jones2006kinematics}, we can derive the curve parameters (i.e., configuration space variables) in terms of the joint variables as
	
	\begin{subequations}
		\begin{align}
			\phi_{j}&=\textstyle \frac{2}{3r_j} \sqrt{ \sum_{i=1}^{3} \textstyle \left(l_{ji}^{2}-l_{ji}\ l_{j \!\!\!\!\!\!\!\mod\left(i,3\right)+1}\right)}\label{eq:cp_phi},\  \mbox{and}\\
			\theta_{j}&=\arctan\left\{ \sqrt{3}\left(l_{j3}-l_{j2}\right),l_{j2}+l_{j3}-2l_{j1}\right\}. 
			\label{eq:cp_theta}
		\end{align}\label{eq:cp2l}
	\end{subequations}%
	
	We derive the homogeneous transformation matrix (HTM) for any \textit{j}-th soft limb, $\mathbf{T}_j\in\mathbb{SE} \left(3\right)$ as
	\begin{align}
		\mathbf{T}_{j}\left(\boldsymbol{q},\xi\right) & =\mathbf{R}_{Z}\left(\theta_{j}\right)\mathbf{P}_{X}\left(\textstyle \frac{L}{\phi_{j}}\right)\mathbf{R}_{Y}\left({\xi} \phi_{j}\right)\mathbf{P}_{X}\left(\textstyle -\frac{L}{\phi_{j}}\right)\mathbf{R}_{Z}\left(-\theta_{j}\right)\nonumber \\
		& =\left[\begin{array}{cc}
			\mathbf{R}_j\left(\boldsymbol{q,{\xi}}\right) & \mathbf{p}_j\left(\boldsymbol{q,{\xi}}\right)\\
			0 & 1
		\end{array}\right],\label{eq:htm}
	\end{align}
	where $\mathbf{R}_Z \in\mathbb{SO} \left(3\right)$ and $\mathbf{R}_Y \in\mathbb{SO} \left(3\right)$ define taskspace rotation matrices about $+Z_j$ and $+Y_j$, respectively, while $\mathbf{P}_X\!\in\mathbb{R}^{3}$ defines the taskspace translation matrix along $+X_j$.  $\mathbf{R}_j \in\mathbb{SO} \left(3\right)$ and $\mathbf{p}_j\!\in\mathbb{R}^{3}$ give the homogeneous rotation and position matrices, respectively. The scalar $\xi\in [0,1] $ 
	defines any point along the neutral axis
	of the limb (Fig. \ref{fig:Fig3_Schematicdiagram}B). Refer to \cite{godage2015modal} for more information about the derivation.
 
	\subsection{Inverse Kinematics of Soft Limbs\label{subsub:Inverse-Kinematic-Model}}

 \begin{figure}[tb] 
		\centering
		\includegraphics[width=1\linewidth]{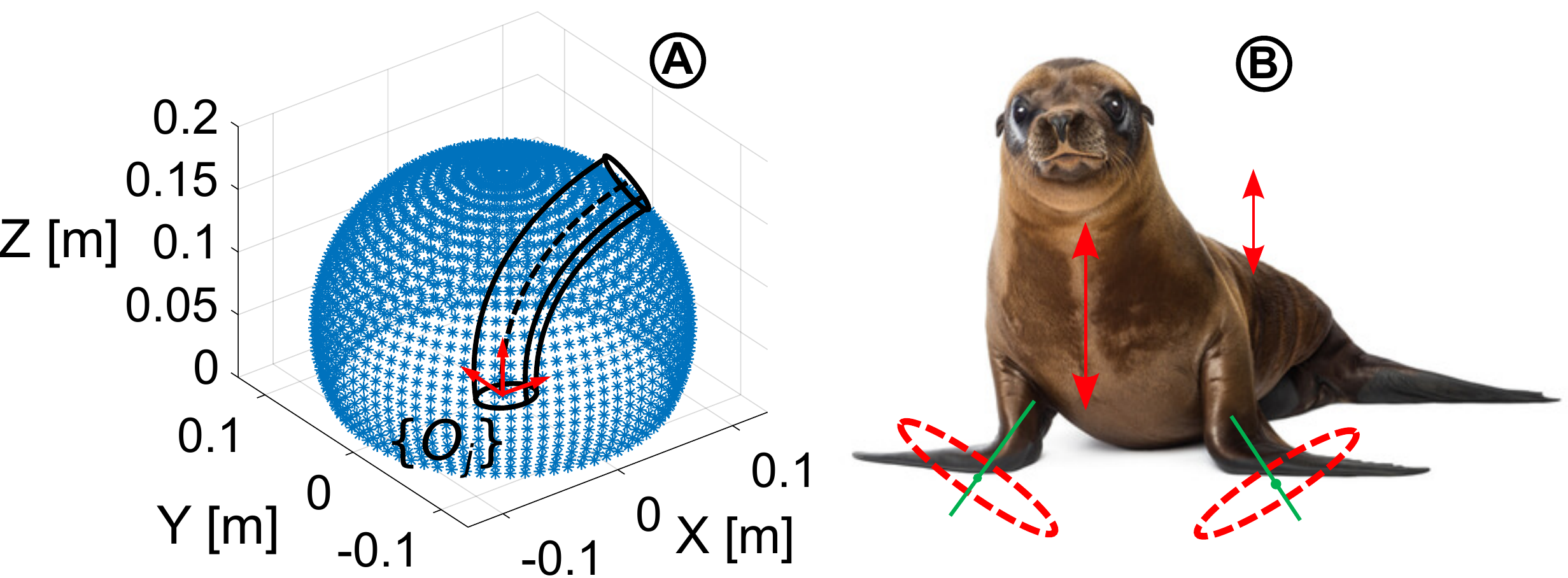}
		\caption{(A) Taskspace of a soft limb in its body coordinate frame, (B) Pinniped terrestrial crawling with limb, body, and head movements.}
		\label{fig:Fig4_LimbTaskspaceBioinspiration} 
	\end{figure}
 
	The relationship between the curve parameters and taskspace coordinates at the tip (i.e., $\xi=1$), $\mathbf{p}_j$ of \eqref{eq:htm}, is given by
	\begin{subequations}
		\begin{align}
			x_{j} & =L\phi_{j}^{-1}\cos\left(\theta_{j}\right)\left\{ 1-\cos\left(\phi_{j}\right)\right\}, \label{eq:x}\\
			y_{j} & =L\phi_{j}^{-1}\sin\left(\theta_{j}\right)\left\{ 1-\cos\left(\phi_{j}\right)\right\},\ \mbox{and}  \label{eq:y}\\
			z_{j} & =L\phi_{j}^{-1}\sin\left(\phi_{j}\right),\label{eq:z}	
		\end{align} \label{eq:xyz_tp}
	\end{subequations}
	$\!\!\!$%
	where $x_{j}$, $y_{j}$, and $z_{j}$ are the position vector elements w.r.t.~the soft limb body coordinates frame, $\{\boldsymbol{O_{j}}\}$.

	A soft limb taskspace -- obtained from the kinematic model in \eqref{eq:htm} -- is a symmetric shell about the $+Z_j$ axis of its body coordinate frame,
	as elucidated, 
	$\{\boldsymbol{O_{j}}\}$, in Fig. \ref{fig:Fig4_LimbTaskspaceBioinspiration}A. 
	Recall that, because of the length constraint imposed by the backbone (Sec. \ref{subsub:Forward-Kinematic-Model}), there are only two kinematic DoFs. Thus we can use two taskspace variables,  $x_j$ and $y_j$, to derive the curve parameters $\theta_{j}$ and $\phi_{j}$. This can be done by solving for 
	\eqref{eq:xy_ik} that maps
 % \todo{Iyad: What maps?}
 taskspace to curve parameters. 	Note that, there is no closed-form solution to \eqref{eq:xy_ik}. Thus, in this work, we utilize MATLAB's `fmincon' constrained optimization routine to solve it. 
	
	\begin{subequations}
		\begin{align}
			\theta_{j} & =\arctan\left(y_{j},x_{j}\right),\label{eq:ik_theta}\\
			\phi_{j}^{-1}\left\{1-\cos\left(\phi_{j}\right)\right\} & =L^{-1}\sqrt{x_{j}^{2}+y_{j}^{2}}.\label{eq:ik_phi}
		\end{align}
		\label{eq:xy_ik}
	\end{subequations}%
	
	\subsection{Complete Robot Kinematics\label{subsub:Robot-Kinematic-Model}}
	
	Refer to the schematic of the robot shown in Fig. \ref{fig:Fig3_Schematicdiagram}C. Utilizing 
	\eqref{eq:htm}, the HTMs of limbs, $\mathbf{T}_{Limb_j}\in\mathbb{SE} \left(3\right)$ relative to the robot coordinates frame, \{$\boldsymbol{O}_{R}$\}, located at the geometric center of the tetrahedral joint can be expressed as
	
	\begin{subequations}
		\begin{align}
			\mathbf{T}_{Limb_1}\left({q{_1}},{\xi}\right) &=\mathbf{T}_{1}\left({q_1},{\xi}\right),\label{eq:Tlimb1}\\
			\mathbf{T}_{Limb_2}\left({q_2},{\xi}\right) &=\mathbf{R}_{Y}\left({\textstyle \frac{\pi}{2}+\delta}\right )\mathbf{R}_{Z}\left({\textstyle {\pi}}\right)\mathbf{T}_{2}\left({q_2},{\xi}\right),\label{eq:Tlimb2}\\
			\mathbf{T}_{Limb_3}\left({q_3},{\xi}\right) &=\mathbf{R}_{Y}\left({\textstyle \frac{\pi}{2}+\delta}\right )\mathbf{R}_{Z}\left({\textstyle \frac{5\pi}{3}}\right )\mathbf{T}_{3}\left({q_3},{\xi}\right),\label{eq:Tlimb3}\\
			\mathbf{T}_{Limb_4}\left({q_4},{\xi}\right) &=\mathbf{R}_{Y}\left({\textstyle \frac{\pi}{2}+\delta}\right )\mathbf{R}_{Z}\left({\textstyle \frac{7\pi}{3}}\right )\mathbf{T}_{4}\left({q_4},{\xi}\right),\label{eq:Tlimb4}
		\end{align} 
		\label{eq:T_robot}
	\end{subequations}
	$\!\!\!\!\!\!$ where $\delta$ is $19.47^{\circ}$ (Fig. \ref{fig:Fig3_Schematicdiagram}C) computed from the tetrahedral geometry. The complete kinematic model of the \textit{j}-th limb of the robot can be obtained utilizing \eqref{eq:T_robot} with a floating-base coordinate frame, $\mathbf{T}_{b}\in\mathbb{SE} \left(3\right)$ as below.
	\begin{align}
		\mathbf{T}_{Limb_j}\left({q}_{b},{q_j},{\xi}\right) &= \mathbf{T}_b\left ( q_{b} \right ) \mathbf{T}_{Limb_j} \left({q_j},{\xi}\right),\label{eq:Tcomplete}\\
		\mathbf{T}_b(q_b) &= \left[\begin{array}{cc}
			\mathbf{R}_Z\left({\alpha}\right)\mathbf{R}_Y\left({\beta}\right)\mathbf{R}_X\left({\gamma}\right) & \mathbf{p}_b\\
			0 & 1
		\end{array}\right].\label{eq:Tb}
	\end{align}
	Herein, ${q}_{b}=\left[x_{b},y_{b},z_{b},\alpha,\beta,\gamma\right]$ with $\left[\alpha,\beta,\gamma\right]$ and $\mathbf{p}_b=[x_{b},y_{b},z_{b}]^\mathsf{T}$ denote the orientation and translation variables of \{$\boldsymbol{O}_{R}$\} relative to the global coordinate frame \{$\boldsymbol{O}$\} (Fig. \ref{fig:Fig3_Schematicdiagram}C). 

	\begin{figure}[tb] 
	\centering
	\includegraphics[width=0.9\linewidth]{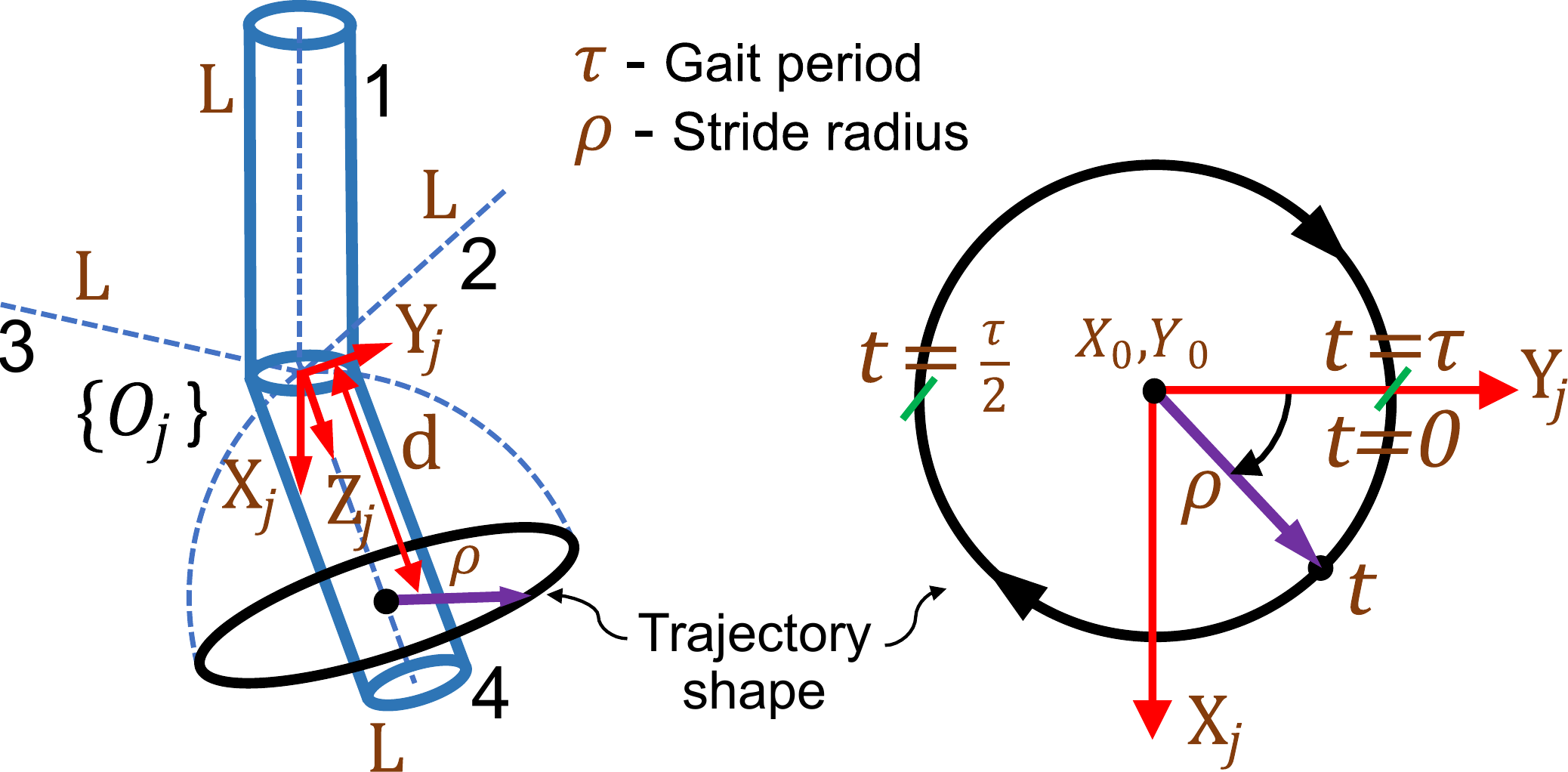}
	\caption{Fundamental motion trajectory of a soft limb.}
	\label{fig:Fig5_TrajectoryGeneration} 
	\end{figure}

	\section{Locomotion Trajectory Generation\label{sec:Trajectory-Generation}}
%	We use limb kinematics in Sec. \ref{subsec:System-Model} to parameterize and derive locomotion gaits as described below. 
	
	\subsection{Fundamental Limb Motion \label{subsec:Limb-Motion}}
	
	The locomotion gaits derived here are inspired by the terrestrial crawling 
	of pinnipeds (Fig. \ref{fig:Fig4_LimbTaskspaceBioinspiration}B).
%	We use the 
	We use limb kinematics in Sec. \ref{subsec:System-Model} to parameterize and derive 
%	locomotion gaits as described below. 
	circular taskspace movement of the limb tip
	as the fundamental limb motion. 
	For any \textit{j}-th soft limb -- shown in Fig. \ref{fig:Fig5_TrajectoryGeneration} --
	we define a circular 
	trajectory 
	of radius $\rho$
	at $d$ distance from the limb's origin, $\{O_j\}$, and period, $\tau$. 
	At time $t$, the tip position
	relative to $\{O_j\}$ is given by 
	\begin{align}
		{x_j=\rho \sin\left(\textstyle -\frac{2\pi t}{\tau}\right),\ y_j=\rho \cos\left(\textstyle -\frac{2\pi t}{\tau}\right),\ z_j=d}
		\label{eq:xyzj}
	\end{align}
	
	We apply uniformly distributed $t \in \left[0,\tau\right]$ values on \eqref{eq:xyzj} to obtain a 100-point taskspace trajectory corresponding to the circular limb motion. 
	We transform the taskspace trajectory to configuration space trajectory using the inverse kinematic model 
	described in Sec. \ref{subsub:Inverse-Kinematic-Model}. Subsequently, \eqref{eq:l2cp} is used to map the configuration space trajectory ($\theta_j,\phi_j$) to 
	the jointspace trajectory ($l_{ji}$). 
%	-- variable length $\left[ {{l_{1}}(t),\,{l_{2}}(t),\,{l_{3}}(t)} \right]^{T}$.
	
	\begin{figure}[tb] 
		\centering
		\includegraphics[width=1\linewidth]{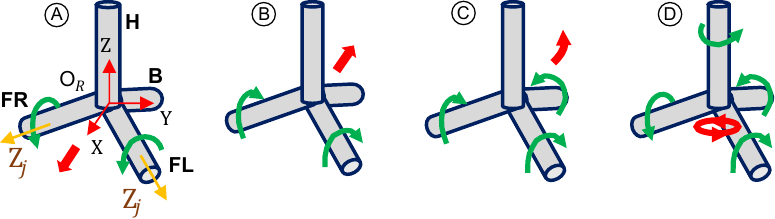}
		\caption{Limb trajectories: (A) Forward crawling, (B) Backward crawling, (C) Crawling-and-turning (Leftward), (D) In-place turning (counterclockwise).}
		\label{fig:Fig6_LimbLevelTrajectoriesandTurning} 
	\end{figure}
 
	\subsection{Effect of Center of Gravity\label{subsec:cog}}
    The center of gravity (CoG) 
%    or the average location 
    of the
%    the weight of the 
    robot helps 
%    track the 
    stabilize
%    ty during 
    locomotion~\cite{wu2013stable}.
%    while locomoting. 
    We compute the robot CoG to investigate and regulate  locomotion stability. 
%    and the crawling thrusts. 
    From \cite{godage2015accurate}, the CoG 
%    position 
%    vector 
    of a limb, $\mathbf{c}_j\in \mathbb{R}^{3}$, relative to its body coordinate frame {$O_j$\} is 
%    can be written as 
    
	\begin{align}
		\mathbf{c}_j(q_j)=\int_{0}^{1}\mathbf{p}_j(\xi,q_j)d\xi. 
		\label{eq:CoG}
	\end{align}
	
	Substituting $\mathbf{p}_j$ in \eqref{eq:xyz_tp} into \eqref{eq:CoG}, $\mathbf{c}_j(q_j)$ can be derived as
	\begin{align}	
		\mathbf{c}_j(q_j)=\frac{L}{\phi_j^2} \begin{bmatrix} \cos \left ({\theta}_j\right )\left ( \phi_j-\sin \left ({\phi}_j\right )\right)
			\\ \sin \left ({\theta}_j\right )\left ( \phi_j-\sin \left ({\phi}_j\right )\right)
			\\\left ( 1-\cos \left ({\phi}_j\right )\right)
		\end{bmatrix}.
		\label{eq:CoG_beta}
	\end{align}
	
	Utilizing the results in \eqref{eq:T_robot} and \eqref{eq:CoG_beta}, CoG relative to the robot coordinate frame, $\{O_R\}$, denoted by $\mathbf{C}_j \in \mathbb{R}^{3}$,
	can be obtained. 
	If the 
%	CoG position 
%%	vector 
%	of the \textit{j}-th limb relative to $\{O_R\}$ is $\mathbf{C}_j \in \mathbb{R}^{3}$, and weight 
	mass of the \textit{j}-th limb is $m_j$, 
	%	and $M$ are 
%	is the 
%	, 
	then CoG 
%	position 
%	vector 
	of the robot relative to $\{O_R\}$, $\mathbf{C}_R \in \mathbb{R}^{3}$, can be written as 
%	$\mathbf{C}_R$
	\begin{align}
		\mathbf{C}_R\left(q_j\right)= \frac{1}{\sum_{j=1}^{4} m_j}\sum_{j=1}^{4} m_j\mathbf{C}_j(q_j). 
		\label{eq:CoG_Robot}
	\end{align}	
	
	\subsection{Forward Crawling\label{subsec:fwd-crawling}}

	\begin{figure}[tb] 
		\centering
		\includegraphics[width=1\linewidth]{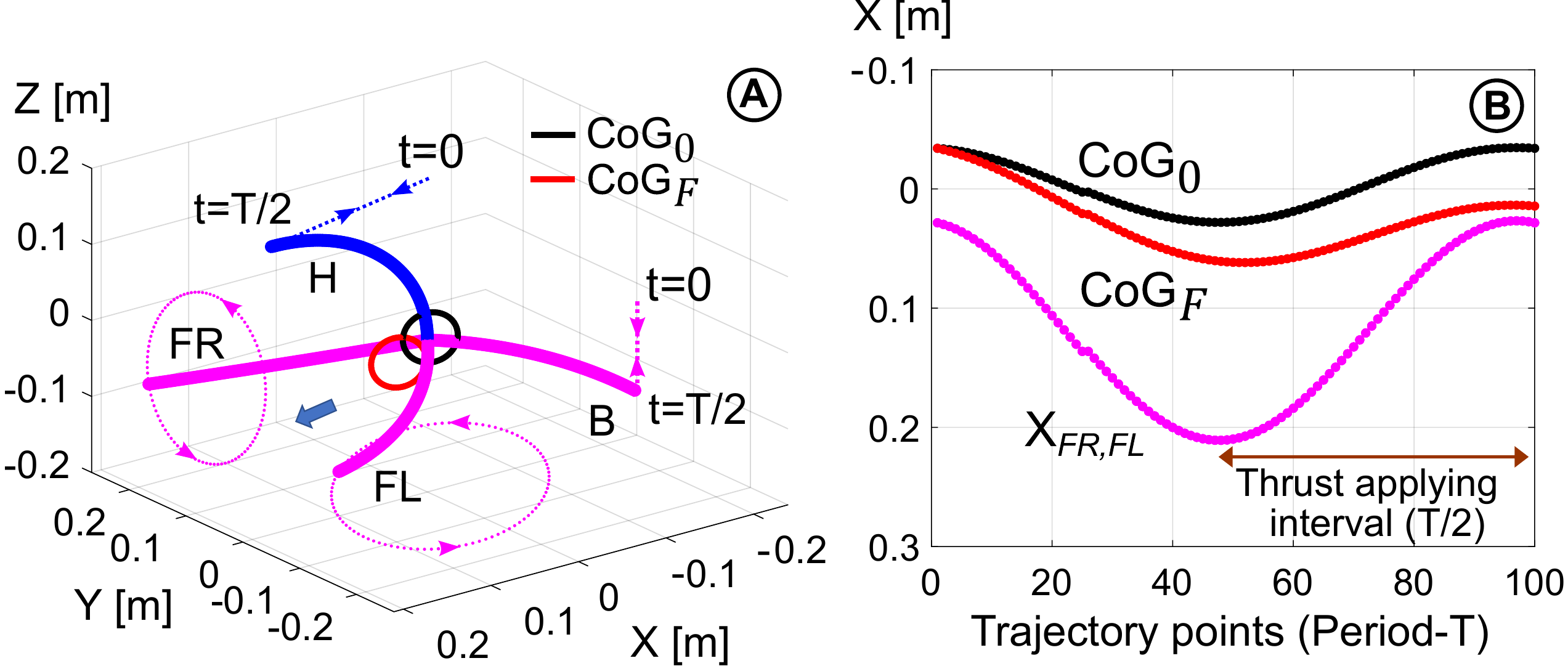}
		\caption{In forward crawling -- (A) Spatial limb displacements and computed CoG trajectories, (B) CoG components and crawling limb tip displacements along the moving direction (i.e., $+X$ axis) relative to $O_R$.}
		\label{fig:Fig7_CoG Trajectory} 
	\end{figure}

    We generate
	forward crawling locomotion
	by simultaneously (i.e., with zero phase offset) replicating
	the limb motion derived in Sec. \ref{subsec:Limb-Motion} in
	FR and FL limbs as illustrated in Fig. \ref{fig:Fig6_LimbLevelTrajectoriesandTurning}A.
    Therein, we move the robot in $+X$ direction, by giving anticlockwise and clockwise motion trajectories to FR and FL limbs w.r.t.~the local coordinate frames thereof, respectively.
	However, achieving forward crawling is challenging as 
 	there is, unlike pinnipeds with their relatively massive bodies, no body 
% forward 
	(or support limb) to counterbalance the angular moment generated by crawling forelimbs. Because of that, forward crawling in the proposed robot can induce instability.

	We circumvent this limitation by 
	controlling the CoG position 
	given by \eqref{eq:CoG_Robot} to obtain a more stable forward crawling gait as described below. 
	Refer to Fig. \ref{fig:Fig7_CoG Trajectory}A for 
	the limb movements and 
	CoG trajectories during a forward crawling cycle. 
	We cyclically 
	and proportionally bend
	the Head (H) limb towards the moving direction from a straight position ($\phi =0^{\circ}$) to 
	a value computed using \eqref{eq:CoG_Robot},
	$\phi =90^{\circ}$, 
	during a locomotion cycle (Fig. \ref{fig:Fig7_CoG Trajectory}A).
	This dynamic CoG control approach stabilizes the movement 
	by counteracting instantaneous torque imbalances. 
	We generate an additional thrust from the Back (B) limb (located on the opposite side) by actuating it in a manner that supports forward propelling.
    Therein, the B limb is gradually bent in a linear trajectory against the moving floor (Fig. \ref{fig:Fig7_CoG Trajectory}A). As a consequence, the resultant limb displacement torque increases. Readers are referred to the experimental video on forward crawling to further understand the above limb actuating mechanism. 
	
	The impact of 
	H and B limb actuation on crawling thrusts can be 
	visualized by tracking the 
	robot CoG and limb movements as shown in Figs. \ref{fig:Fig7_CoG Trajectory}A and \ref{fig:Fig7_CoG Trajectory}B. The CoG$_0$ denotes the CoG trajectory when H and B limbs are not actuated. When they are actuated, CoG$_0$ shifts towards the moving direction ($+X$) as noted by CoG$_F$ in Fig. \ref{fig:Fig7_CoG Trajectory}A. Figure \ref{fig:Fig7_CoG Trajectory}B shows computed CoG$_0$, CoG$_F$, and crawling limb tips (X$_{FR}$, X$_{FL}$) in the moving direction relative to $O_R$. During the crawling thrust applying interval (i.e., ground contact period), the robot CoG converges and closely follows crawling limb tips as noted by CoG$_F$ in Fig. \ref{fig:Fig7_CoG Trajectory}B. 
	It causes 
%	to 
	an increase in the weight-induced torque 
	supported by the crawling limbs (FR \& FL).
%	that make the ground contact. 
	As a consequence, with the increase in ground-limb reaction forces, the 
%	applied 
	crawling thrusts increase.

	\subsection{Backward Crawling\label{subsec:bwd-crawling}} 

    The backward crawling is referred to as moving in the $-X$ direction (Fig. \ref{fig:Fig1_IntroductionImage}B).
    Here, the limb motion derived in Sec. \ref{subsec:Limb-Motion} is simultaneously applied to FR and FL limbs in the opposite direction to that of the forward crawling, i.e., FR and FL limbs are given clockwise and anticlockwise motion trajectories respectively, as illustrated in Fig. \ref{fig:Fig6_LimbLevelTrajectoriesandTurning}B.
	We keep the Head (H) limb bent 
	in the $-X$ direction (i.e., backward) for shifting the robot CoG toward FR and FL limbs for improved stability and generating more thrust from the increased weight (reaction forces) at the limbs-ground contact \cite{wang2021design}. 
	Concurrently, the Back (B) limb is 
	bent upward (in the $+Z$ direction of $O_R$) to reduce the contact surface and minimize the frictional resistance \cite{wang2021design}. 
     %Above limb actuation can be further understood by examining Figs. \ref{fig:Fig8_SpaceTrajectoriesFwdCrawling}A, \ref{fig:Fig8_SpaceTrajectoriesFwdCrawling}B, and \ref{fig:Fig8_SpaceTrajectoriesFwdCrawling}C, that show taskspace, configuration space, and jointspace trajectories of four limbs, respectively, relative to limbs' body coordinate frames for $\tau=4~s$ and $\rho=0.10~m$.

 %    \begin{figure}[tb] 
	% 	\centering
	% 	\includegraphics[width=0.99\linewidth]{figures/Fig8_SpaceTrajectoriesFwdCrawling.pdf}
	% 	\caption{In backward crawling ($\tau=4~s,\ \rho=0.10~m$), (A) Taskspace, (B) Configuration space, (C) Variable-length jointspace trajectories of four limbs relative to individual limb coordinate frames.}
	% 	\label{fig:Fig8_SpaceTrajectoriesFwdCrawling} 
	% \end{figure}
 
	\subsection{Crawling-and-Turning \label{subsec:crawling-turning}}

    Pinnipeds use peristaltic body movement to propel forward since the bulk of the body weight is distributed towards the back (body)~\cite{kuhn2012walking}. But, the proposed soft robot design has a symmetric weight distribution and thus it is difficult to maintain stability while propelling forward. As a consequence, the robot shows limited frontal movements. Conversely, when propelling backward, the torque imbalance is countered by the Body (i.e., B limb). It enables the use of the B limb in turning only in backward movements. Therefore, we opt to achieve turning in the backward direction. 
%   
%	For instance, 
	To achieve turning locomotion, 
%	(Fig. \ref{fig:Fig6_LimbLevelTrajectoriesandTurning}C), 
	we additionally actuate the B limb similarly to 
	straight crawling limbs (FR \& FL) discussed in Sec. \ref{subsec:fwd-crawling}.
%    
%    By applying an additional angular torque through the B limb, we can change the moving direction of the robot. 
    For example, a clockwise trajectory of the B limb results in a leftward turn (Fig. \ref{fig:Fig6_LimbLevelTrajectoriesandTurning}C), while changing the direction of the B limb to anticlockwise results in a rightward turn.
%        
%    The idea is to change the moving direction of the robot by applying
%	an additional angular torque 
%	via the Body (B) limb.
%    A leftward turning is achieved by providing a clockwise trajectory to the B limb (Fig. \ref{fig:Fig6_LimbLevelTrajectoriesandTurning}C). A rightward turning is achieved by changing the turn direction of the B limb to anticlockwise. %
	We replicate the B limb motion with different stride radii to control the turning effect. For example, a relatively large stride trajectory of the B limb can turn the robot efficiently (see results in Sec.~\ref{sec:Experimental-Validation} and experimental videos).
	
	\subsection{In-place Turning \label{subsec:in-placeTurning}}
	
	In-place turning is referred to as the rotation about the robot $+Z$ axis (Fig. \ref{fig:Fig1_IntroductionImage}A). It is achieved by crawling all ground-contacting limbs in the same direction of rotation (clockwise\slash counterclockwise) as shown in Fig. \ref{fig:Fig6_LimbLevelTrajectoriesandTurning}D. Additionally, we actuate the Head (H) limb in the same direction of rotation in a circular trajectory at 
	the same
%	an identical 
	angular velocity. In that way, we shift the CoG of the Head (H) limb into the direction of rotation and support the turning. We can reverse the direction of in-place turning by reversing the direction of crawling in all limbs. 

    \begin{figure}[tb] 
		\centering
		\includegraphics[width=0.95\linewidth]{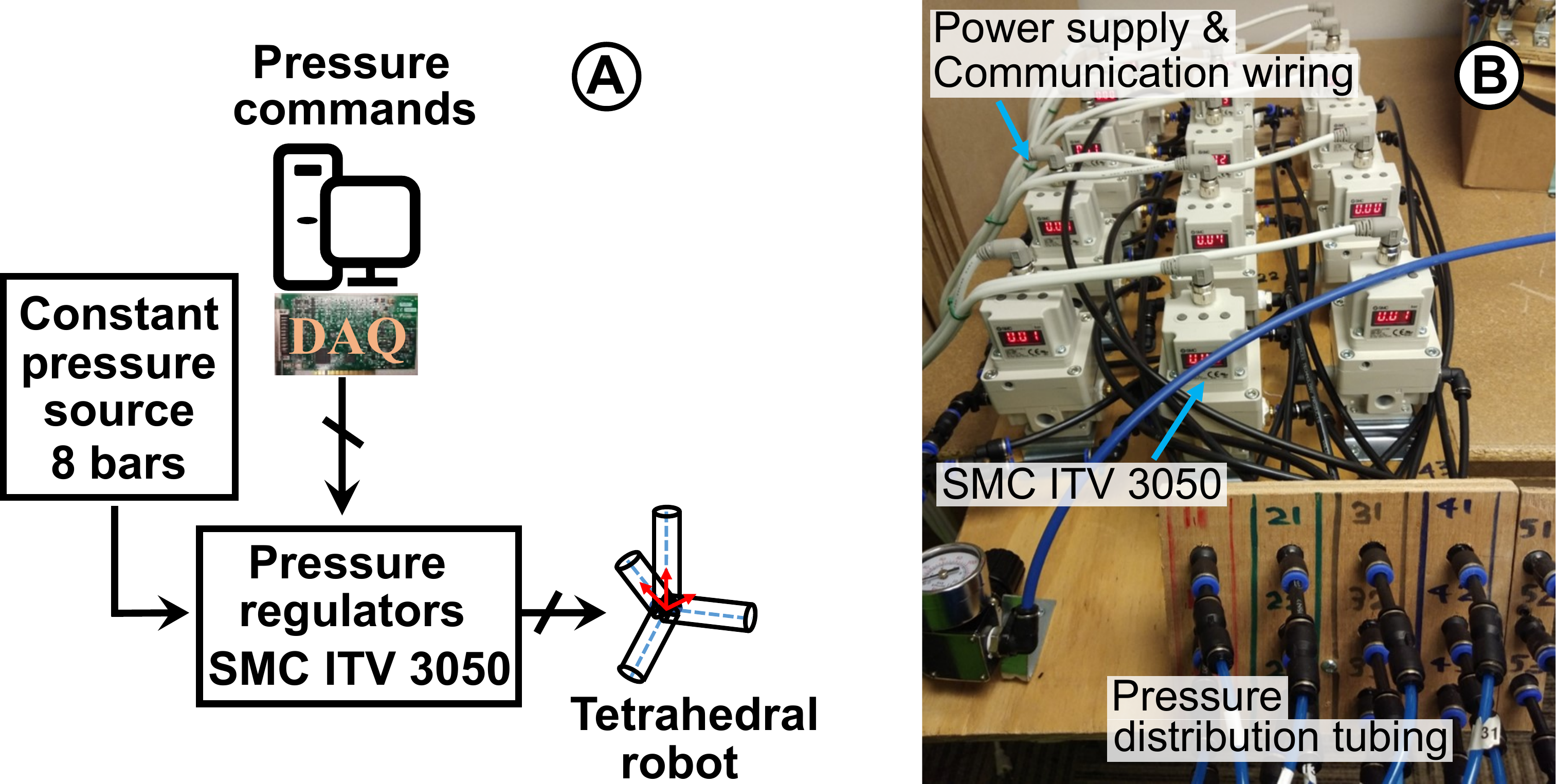}
		\caption{(A) Robot actuation setup, (B) Pressure regulator assembly.}
		\label{fig:Fig9_ActuationSetup}
	\end{figure} 

	\begin{figure*}[tb] 
		\centering
		\includegraphics[width=0.995\textwidth, height=0.45\textwidth]{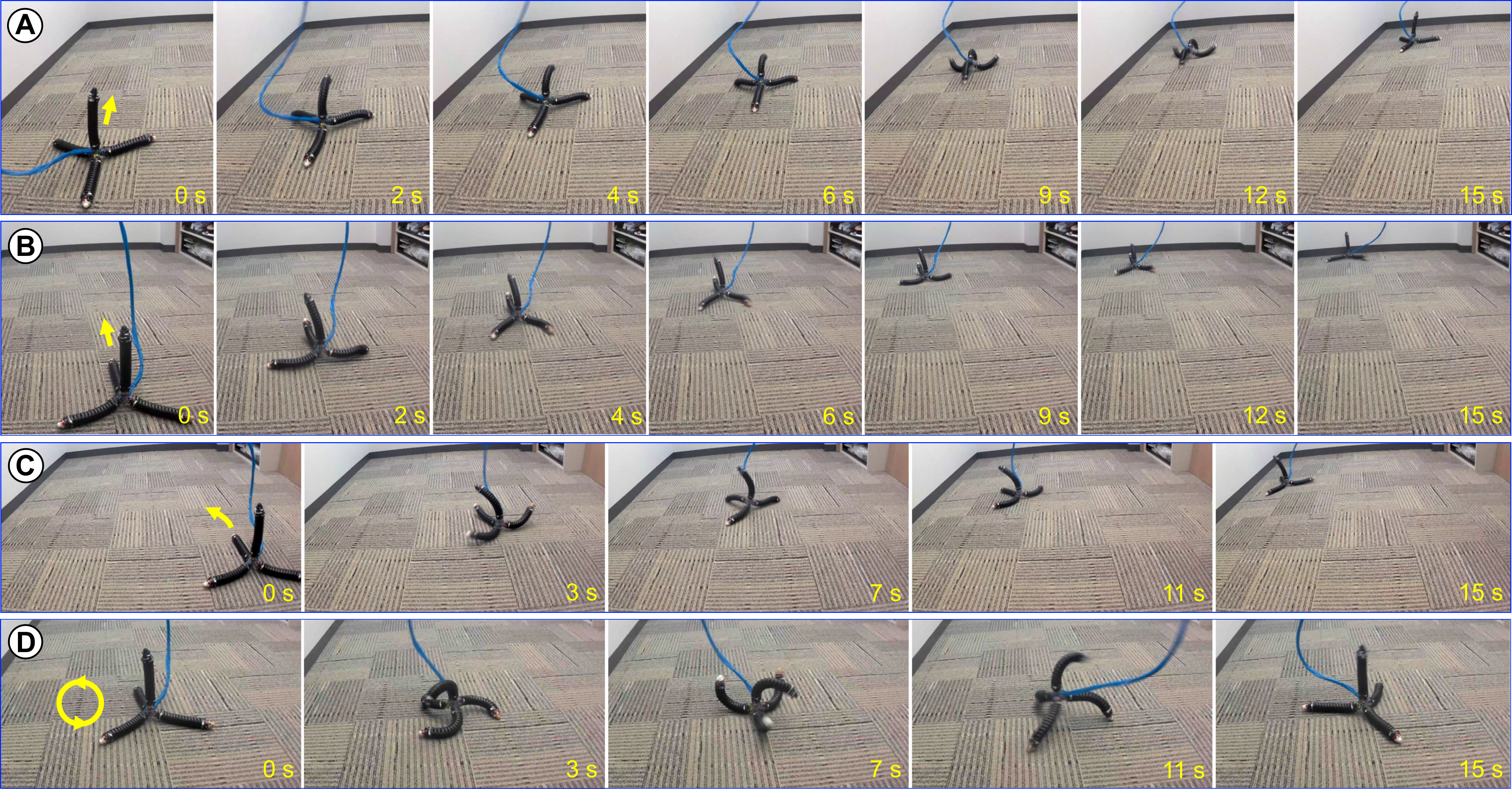}
		\caption{(A) Forward crawling, (B) Backward crawling, (C) Crawling-and-turning (leftward), (D) In-place turning (counterclockwise), at $0.10~m-1.00~Hz$.}
		\label{fig:Fig10_CrawlingStraightandTurning} 
	\end{figure*}

	\section{Experimental Validation\label{sec:Experimental-Validation}}
	
	\subsection{Experimental Setup\label{subsec:Experimental-Setup}}
	
	The experimental setup for the robot is depicted in Fig. \ref{fig:Fig9_ActuationSetup}A. Air pressure is supplied from an $8-bar$ pneumatic source to digital proportional pressure regulators (ITV3050 31F3N3, SMC USA), as shown in Fig. \ref{fig:Fig9_ActuationSetup}B. The pressure regulators receive commands from a MATLAB Simulink Desktop Real-Time model through a data acquisition card (PCI-6703, NI USA), which sends an $0-10~V$ analog voltage signal. To make the soft limbs move and perform locomotion, the jointspace trajectories obtained in Sec. \ref{sec:Trajectory-Generation} 
 %(such as those in Fig. \ref{fig:Fig8_SpaceTrajectoriesFwdCrawling}C) 
 must be converted into actuation pressure trajectories and input into the actuation setup shown in Fig. \ref{fig:Fig9_ActuationSetup}. The jointspace-pressure mapping reported in~\cite{arachchige2021soft} is used to generate the corresponding pressure inputs. The robot is tested on a carpeted floor with approximately uniform friction, as seen in Fig. \ref{fig:Fig10_CrawlingStraightandTurning}.
	
%	The robot experimental setup is shown in Fig. \ref{fig:Fig9_ActuationSetup}A. An 8 $bar$ pneumatic source supplies air pressure to digital proportional pressure regulators (ITV3050 31F3N3, SMC USA -- Fig. \ref{fig:Fig9_ActuationSetup}B). A data acquisition card (PCI-6703, NI USA) is interfaced with a MATLAB Simulink Desktop Real-Time model to send pressure commands to pressure regulators via a $0-10~V$ voltage signal. Note that, to actuate soft limbs and achieve locomotion, the jointspace trajectories (i.e., length changes of PMAs) obtained in Sec.\ref{sec:Trajectory-Generation} (such as those in Fig. \ref{fig:Fig8_SpaceTrajectoriesFwdCrawling}C) should be converted into actuation pressure trajectories and input them via the actuation setup in Fig. \ref{fig:Fig9_ActuationSetup}. We adopted the jointspace -- pressure mapping reported in \cite{arachchige2021soft} to generate corresponding pressure inputs. The robot is tested on a carpeted floor (with approximately uniform friction), as shown in Fig. \ref{fig:Fig10_CrawlingStraightandTurning}.
	
	\subsection{Testing Methodology \label{subsec:Testing-Methodology}}
	
	We actuated each gait for $15~s$ with a $3~bar$ actuator 
%	supply 
	pressure ceiling (based on PMAs' ability to achieve the required limb deformation). The frequency range, $\{0.75, 1.00, 1.25\}~Hz$ was chosen based on the operational bandwidth of PMAs to replicate meaningful locomotion. With 03 frequency combinations, we conducted 54 experiments for 06 straight crawling gaits, 06 crawling-and-turning gaits, and 06 in-place turning gaits as detailed in Secs. \ref{subsec:Testing-Straight-Crawling-Turning} and \ref{subsec:TestingTurning}.

	\subsection{Forward and Backward Crawling Gaits }\label{subsec:Testing-Straight-Crawling-Turning}

	We generated a total of 18 combinations of forward and backward crawling locomotion trajectories, with three gaits in each direction, using three different stride radii ($\rho_{1}=0.06~m$, $\rho_{2}=0.08~m$, $\rho_{3}=0.10~m$). Figures \ref{fig:Fig10_CrawlingStraightandTurning}A and \ref{fig:Fig10_CrawlingStraightandTurning}B show the progression of the robot during forward and backward crawling at the $0.10~m - 1.00~Hz$ stride radius-frequency combination. Complete videos of the experiments are included in our multimedia submission. To determine the robot's moving distance along the $X$ and $Y$ directions, we used the perspective image projection approach reported in~\cite{arachchige2021soft,arachchige2023dynamic}. This approach utilized video feedback and floor carpet geometry data to estimate the distances. Note that some deviation from the intended gait is expected due to the performance variations of the custom-built PMAs powering the soft limbs.

	%%	In this work, w
	%	We generated 18 combinations of forward and backward crawling locomotion trajectories (03 gaits in each direction) for 03 stride radii ($\rho_{1}=0.06~m$, $\rho_{2}=0.08~m$, $\rho_{3}=0.10~m$). Figures \ref{fig:Fig10_CrawlingStraightandTurning}A and \ref{fig:Fig10_CrawlingStraightandTurning}B show the progression of the robot during forward and backward crawling at $0.10~m-1.00~Hz$ stride radius$-$frequency combination. Our multimedia submission shows complete videos of the experiments. We obtained the robot's moving distance along the $X$, $Y$ directions using the perspective image projection approach reported in \cite{arachchige2021soft}. Therein, video feedback and 
	%%	available 
	%	floor carpet geometry data were used for distance estimations. Note that gait divergence is expected in all gaits during experiments due to performance variation of custom-built PMAs powering the soft limbs. 
	We present the performance of each crawling gait in terms of estimated robot speed, which is shown in Table~\ref{Table:Performancesummary-crawling}. The experiments revealed that the robot achieved higher speeds at larger stride radii ($0.10~m$) and moderate actuation frequencies ($1.00~Hz$). This is because larger crawling strides generate stronger limb displacement torques on the floor than 
%	while 
	smaller strides.
% 	produce weaker torques. 
 	In addition, moderate actuation frequencies enable air pressure to reach the PMAs in a timely manner through the long pneumatic tubes, allowing for desired limb deformation without distortion of the torque amplitude. The highest recorded crawling speed was $16.9~cm/s$ ($0.65$ body length/second), which is a $38$-fold increase from the state-of-the-art reported in~\cite{wang2021design}, $0.37 cm/s$. 
% 	($0.017$ body length/second).
	
%	We present the performance of each crawling gait in terms of the estimated robot speed shown in Table \ref{Table:Performancesummary-crawling}. The robot experimentally recorded large speeds at higher stride radii ($0.10~m$) and moderate actuation frequencies ($1.00~Hz$). This is because large crawling strides generate higher limb displacement torques on the floor and vice versa for small strides. On the other hand, moderate actuation frequencies allow air pressure to reach PMAs in due time along long pneumatic tubes; hence desired limb deformation is achieved. As a consequence, the torque amplitude is not distorted. The recorded maximum crawling speed, $16.9~cm^{-1}$ ($0.65~\text{body~length}/\text{second}$) is a 38-fold speed increase than that of the state-of-the-art in \cite{wang2021design}, $0.37~cm^{-1}$ ($0.017~\text{body~length}/\text{second}$).

	In forward crawling, the robot must perform additional limb deformations, as described in Sec.~\ref{subsec:bwd-crawling}, in order to maintain balance and generate additional forward propulsion. As a result, 
%	the 
	forward crawling recorded lower speeds compared to backward crawling at all times. The accompanying video further demonstrates that although forward crawling resembles pinniped locomotion, it is less efficient in maintaining forward locomotion stability.
	
%	In forward crawling, the robot has to execute additional limb deformations (described in Sec. \ref{subsec:bwd-crawling}) to not only balance the robot but also to generate an additional propelling thrust in the moving direction. Therefore, forward crawling experimentally recorded a lower speed than backward crawling at all times. The accompanying video provides further evidence that though forward crawling mimics pinniped locomotion, it is less efficient in maintaining moving stability. 

\begin{table}[tb]
	\setlength{\tabcolsep}{3.0pt} %
	\centering
	\caption{\textsc{Performance of Straight Crawling Gaits.}}
	\label{Table:Performancesummary-crawling}
	\begin{tabular}{|c|c|c|c|c|c|c|c|c|c|} 
		\hline
		\multirow{5}{*}{\textbf{Straight Gait}} & \multicolumn{9}{c|}{\textbf{Stride radius of crawling limbs (FR \& FL)}} \\ 
		\cline{2-10}
		& \multicolumn{3}{c|}{$\rho_{1}$ (0.06~$m$)} & \multicolumn{3}{c|}{$\rho_{2}$ (0.08~$m$)} & \multicolumn{3}{c|}{$\rho_{3}$ (0.10~$m$)} \\ 
		\cline{2-10}
		& \multicolumn{3}{c|}{\textbf{Freq.}~[$Hz$]} & \multicolumn{3}{c|}{\textbf{Freq.}~[$Hz$]} & \multicolumn{3}{c|}{\textbf{Freq.}~[$Hz$]} \\ 
		\cline{2-10}
		& 0.75 & 1.00 & 1.25 & 0.75 & 1.00 & 1.25 & 0.75 & 1.00 & 1.25 \\ 
		\cline{2-10}
		& \multicolumn{9}{c|}{\textbf{Mean speed} $[cm/ s]$} \\ 
		\hline
		\textbf{Fwd Crawling} & 5.34 & 7.57 & 7.21 & 7.53 & 10.2 & 9.81 & 9.41 & 11.9 & 10.9 \\
		\hline
		\textbf{Bwd Crawling} & 7.21 & 10.1 & 9.83 & 9.52 & 13.5 & 13.0 & 11.9 & 16.9 & 16.1 \\ 
		\hline
	\end{tabular}
\end{table}

\begin{table}[tb]
	\setlength{\tabcolsep}{2.55pt} %
	\centering
	\caption{\textsc{Performance of Crawling-and-Turning gaits.}}
	\label{Table:Performancesummary-turning}
	\begin{tabular}{|c|c|c|c|c|c|c|c|c|c|}
		\hline
		\multirow{5}{*}{\textbf{Turning Gait}} & \multicolumn{9}{c|}{\textbf{Stride radius of turning limb (B limb)}} \\ 
		\cline{2-10}
		& \multicolumn{3}{c|}{$\rho_{1}$ (0.04~$m$)} & \multicolumn{3}{c|}{$\rho_{2}$ (0.06~$m$)} & \multicolumn{3}{c|}{$\rho_{3}$ (0.08~$m$)} \\ 
		\cline{2-10}
		& \multicolumn{3}{c|}{\textbf{Freq.~[$Hz$]}} & \multicolumn{3}{c|}{\textbf{Freq.~[$Hz$]}} & \multicolumn{3}{c|}{\textbf{Freq.~[$Hz$]}} \\ 
		\cline{2-10}
		& 0.75 & 1.00 & 1.25 & 0.75 & 1.00 & 1.25 & 0.75 & 1.00 & 1.25 \\ 
		\cline{2-10}
		& \multicolumn{9}{c|}{\textbf{Angular speed per unit distance}~$[rad/ (ms)]$} \\ 
		\hline
		\textbf{Leftward Turn} & 1.13 & 1.72 & 1.70 & 1.59 & 2.24 & 2.06 & 2.22 & 2.89 & 2.41 \\ 
		\hline
		\textbf{Rightward Turn} & 1.15 & 1.68 & 1.65 & 1.62 & 2.31 & 2.11 & 2.35 & 2.92 & 2.49 \\
		\hline
	\end{tabular}
\end{table}
 
	\subsection{Turning Gaits\label{subsec:TestingTurning}}

We have successfully generated crawling-and-turning gaits for backward crawling locomotion (as described in Sec.~\ref{subsec:crawling-turning}). We created 03 leftward and 03 rightward turning trajectories by varying the stride radius of the B limb at values of ($\rho_{1}=0.04~m$, $\rho_{2}=0.06~m$, $\rho_{3}=0.08~m$). For these gaits, the FR and FL limbs were actuated at a fixed stride radius of $0.10~m$.

\begin{table}[b]
	\setlength{\tabcolsep}{2.40pt} %
	\centering
	\caption{\textsc{Performance of In-place Turning gaits.}}
	\label{Table:Performancesummary-inplaceturning}
	\begin{tabular}{|c|c|c|c|c|c|c|c|c|c|}
		\hline
		\multirow{5}{*}{\textbf{Turning Gait}} & \multicolumn{9}{c|}{\textbf{Stride radius of crawling limbs (FR, FL, B)}} \\ 
		\cline{2-10}
		& \multicolumn{3}{c|}{$\rho_{1}$ (0.06~$m$)} & \multicolumn{3}{c|}{$\rho_{2}$ (0.08~$m$)} & \multicolumn{3}{c|}{$\rho_{3}$ (0.10~$m$)} \\ 
		\cline{2-10}
		& \multicolumn{3}{c|}{\textbf{Freq.~[$Hz$]}} & \multicolumn{3}{c|}{\textbf{Freq.~[$Hz$]}} & \multicolumn{3}{c|}{\textbf{Freq.~[$Hz$]}} \\ 
		\cline{2-10}
		& 0.75 & 1.00 & 1.25 & 0.75 & 1.00 & 1.25 & 0.75 & 1.00 & 1.25 \\ 
		\cline{2-10}
		& \multicolumn{9}{c|}{\textbf{Angular speed}~$[rad/s]$} \\ 
		\hline
		\textbf{Clockwise} & 2.75 & 3.29 & 3.09 & 3.01 & 3.55 & 3.41 & 3.35 & 3.76 & 3.51 \\ 
		\hline
		\textbf{Counterclockwise} & 2.81 & 3.35 & 3.12 & 3.08 & 3.69 & 3.53 & 3.42 & 3.82 & 3.59 \\
		\hline
	\end{tabular}
\end{table}

For in-place turning, we produced six trajectories to represent clockwise/counterclockwise turning under three stride radii ($\rho_{1}=0.06~m$, $\rho_{2}=0.08~m$, $\rho_{3}=0.10~m$). During these gaits, all limbs, including the Head (H) limb, were actuated under the same stride radii as each crawling gait. Figures~\ref{fig:Fig10_CrawlingStraightandTurning}C and \ref{fig:Fig10_CrawlingStraightandTurning}D show the leftward crawling-and-turning gait and counterclockwise in-place turning gait, respectively. The performance of these trajectories is presented in Tables~\ref{Table:Performancesummary-turning} and \ref{Table:Performancesummary-inplaceturning}, respectively.
We experimentally measured the turn angle and $X-Y$ floor displacement for all gaits using the method described in the straight crawling in Sec.~\ref{subsec:Testing-Straight-Crawling-Turning}. We then calculated the angular speed per unit distance for crawling-and-turning gait and the angular speed for in-place turning gait. According to the data in Table~\ref{Table:Performancesummary-turning}, the effectiveness of turning increases with the stride radius of the turning limb. Similarly, the data in Table~\ref{Table:Performancesummary-inplaceturning} indicates that the robot performs well in replicating in-place turning at higher stride radii, due to the increase in relative turn displacement torque with the applied trajectory stride radius.

	\section{Conclusions \label{sec:ConclusionFuture Work}}

The soft-limbed robots have great potential for use in locomotion applications. We have designed a soft-limbed robot, which mimics pinniped locomotion, with a tetrahedral topology. The modular design approach for developing the robot was explained. Forward and inverse kinematic models for a single soft limb, as well as a complete floating-base kinematic model for the entire robot, were derived. The task-space trajectories for fundamental limb motion were proposed, and joint-space trajectories were obtained using kinematic models for forward/backward crawling, crawling-and-turning, and in-place turning gaits. The performance of the pinniped robot was experimentally validated under different stride radii and actuation frequencies, and the results show that the proposed locomotion trajectories were replicated well. Further work will focus on the development of dynamic gaits and closed-loop control of pinniped locomotion.
	
%    Soft-limbed robots show great potential for usage in locomotive applications. 	
%    We proposed a soft-limbed robot assembled in a tetrahedral topology that 
%	replicates pinniped locomotion.  
%	We detailed the modular design approach for developing the robot.
%	We derived curve parametric forward and inverse kinematic models for a soft limb and a complete floating-base kinematic model for the robot. We proposed taskspace trajectories and used kinematic models to derive jointspace trajectories for 
%%	straight 
%	forward\slash backward crawling, crawling-and-turning, and in-place turning gaits. We experimentally validated the robot under different stride radii and actuation frequencies. Experimental results showed that the pinniped robot can replicate proposed locomotion trajectories well. 
%%	\todo{Iyad: How about concluding with some open problems for future work?}
%	Future work will focus on the dynamic 
%%	locomotion 
%	gaits and closed-loop control of the pinniped locomotion. 
	
	\bibliographystyle{IEEEtran}
	\bibliography{refs}
	
\end{document}